\documentclass[11pt]{article}

\usepackage{mathtools}
\usepackage[round]{natbib}
\usepackage[english]{babel}
\usepackage{hyperref}
\usepackage{authblk}
\usepackage{algpseudocode}
\usepackage{algorithm}

\usepackage{amsmath}
\usepackage{amsfonts}
\usepackage{color}
\usepackage[colorinlistoftodos, backgroundcolor=white, linecolor=gray, textsize=tiny,textwidth=100pt]{todonotes}
\definecolor{db}{rgb}{0, 0, 0.7}

\title{The configurable tree graph (CT-graph): measurable problems in partially observable and distal reward environments for lifelong reinforcement learning}
\author[1]{Andrea Soltoggio}
\author[1]{Eseoghene Ben-Iwhiwhu}
\author[1]{Christos Peridis}
\author[2]{Pawel Ladosz}
\author[1]{Jeffery Dick}
\author[3]{Praveen K. Pilly}
\author[4]{Soheil Kolouri}
{\small
\affil[1]{Loughborough University, UK\\
emails:\{a.soltoggio,e.ben-iwhiwhu,c.peridis,j.dick\}@lboro.ac.uk}
\affil[2]{UNIST, Republic of Korea,
email: pladosz@unist.ac.kr}
\affil[3]{HRL Laboratories, USA\\
email: pkpilly@hrl.com}
\affil[4]{Vanderbilt University, USA\\
email: soheil.kolouri@vanderbilt.edu}
}
\date{}

\begin{document}
\maketitle
\vspace{-1cm}
\begin{abstract}
This paper introduces a set of formally defined and transparent problems for reinforcement learning algorithms with the following characteristics: (1) variable degrees of observability (non-Markov observations), (2) distal and sparse rewards, (3) variable and hierarchical reward structure, (4) multiple-task generation, (5) variable problem complexity. The environment provides 1D or 2D categorical observations, and takes actions as input. The core structure of the CT-graph is a multi-branch tree graph with arbitrary branching factor, depth, and observation sets that can be varied to increase the dimensions of the problem in a controllable and measurable way. Two main categories of states, decision states and wait states,  are devised to create a hierarchy of importance among observations, typical of real-world problems. A large observation set can produce a vast set of histories that impairs memory-augmented agents.  Variable reward functions allow for the easy creation of multiple tasks and the ability of an agent to efficiently adapt in dynamic scenarios where tasks with controllable degrees of similarities are presented. Challenging  complexity levels can be easily achieved due to the exponential growth of the graph. The problem formulation and accompanying code provide a fast, transparent, and mathematically defined set of configurable tests to compare the performance of reinforcement learning algorithms, in particular in lifelong learning settings. 
\end{abstract}

\section{Introduction}

Many real-world problems are characterized by a large number of observations, confounding and spurious correlations, partially observable states, and distal, dynamic rewards with hierarchical reward structures. Such conditions make it hard for both animal and machines to learn complex skills. The learning process requires discovering what is important and what can be ignored, how the reward function is structured, and how to reuse knowledge across different tasks that share common properties. For these reasons, the application of standard reinforcement learning (RL) algorithms \citep{sutton2018reinforcement} to solve structured problems is often not effective. Limitations of current RL algorithms include the problem of exploration with sparse rewards \citep{pathak2017curiosity}, dealing with partially observable Markov decision problems (POMDP) \citep{ladosz2021deep}, coping with large amounts of confounding stimuli \citep{thrun2000monte,kim2019curiosity}, and reusing skills for efficiently learning multiple task in a lifelong learning setting \citep{mendez2020lifelong}. 

Standard reinforcement learning algorithms are best suited when the problem can be formulated as a single-task problem in observable Markov decision problem (MDP). Under these assumptions, with complete observability and with static and frequent rewards, deep reinforcement learning (DRL) \citep{mnih2015human,li2017deep} has gained popularity due to the ability to learn an approximated Q-value function directly from raw pixel data in the Atari 2600 platform. This and similar algorithms stack multiple frames to derive states of an MDP, and use a basic $\epsilon$-greedy exploration policy. In more complex cases with partial observability and sparse rewards, extensions have been proposed to include more advanced exploration techniques \citep{ladosz2022exploration}, e.g.\ \citep{pathak2017curiosity,burda2018exploration,ecoffet2019go}, and memory systems \citep{hausknecht2015deep,heess2015memory,parisotto2017neural}.  

The need to test RL algorithms in more challenging problems  has led the community to look for increasingly more complex benchmarks \citep{justesen2019deep}. Often, first person view (FPV) games are used to test the ability of RL to cope with partial observability, while games where rewards occur rarely, e.g., Atari Montezuma, are used to test advanced exploration techniques \citep{ecoffet2019go}. However, videogame-based benchmarks have several limitations. Their degree of observability is often unknown or hard to assess. The degree of sparsity and distance of rewards, and their search space might not be clearly defined or measurable. Games might also not be easily configured to express variations of tasks or difficulty, thus limiting tests to a fixed problem complexity and static conditions. Finally, many games are computationally expensive because they generate complex visual fields without necessarily requiring rich policies. As a consequence, algorithms can be assessed only by testing them across a large range of such games, thus requiring considerable computational effort and yet not providing a mathematically defined metric of performance or statistical significance. Moreover, because  the underlying MDP is unknown, it is also unclear how performance of an RL algorithm in a suit of games maps to other real-world problems. Unfortunately, the need to test RL algorithms on a large set of computationally expensive benchmarks slows development and significantly increases costs. Often, only large research groups with abundant computational resources can convincingly show the strength of their algorithms. Recently developed benchmarks such as Minigrid \citep{chevalier2018minimalistic}, ProcGen \citep{cobbe2020leveraging} and Minihack \citep{samvelyan2021minihack} address some of these concerns, offering highly configurable, fast and procedurally generated scenarios. While such benchmarks offer increasingly more flexible and powerful RL benchmarks, they do not have fully measurable search spaces, episode length and reward sparsity. 

This paper introduces a mathematically defined and configurable environment. The environment allows for precisely defined metrics and measurements such as: the degree of partial observability, measurements of distal and sparse rewards, the size of the search space, a defined hierarchy of skills, and variable reward functions.

\begin{table}
\scriptsize
\centering
\begin{tabular}{|l|c|c|}
\hline
\textbf{Property} &\textbf{CT-graph} & \textbf{Videogame benchmarks}\\\hline
Multiple tasks&tasks can be randomly generated&fixed or hard to define/select\\
Task similarity & defined and measurable & not defined/measurable\\
Length of episode & configurable & often unknown\\
Input & configurable data set & defined by the game\\
Output & configurable number of actions & defined by the game\\
Observability & configurable MDP to POMDP & often unknown\\
Sparsity of reward & configurable & often predefined and fixed\\
Computational cost & low & often large/unrelated to complexity\\
Size of search space & configurable and known&often fixed and unknown\\
Optimal policy & configurable and known&often unknown\\
\hline
\end{tabular}
\caption{List of properties that were built in the CT-graph to address the limitations of video game benchmarks.}
\label{tab.1}
\end{table} 

The proposed environment is an abstraction of a decision process informed by visual stimuli and simulated with a configurable tree graph with a set of configurable parameters. At the core of the system is a configurable tree graph that can be expanded both in the depth (i.e.\ the length of the sequence of actions to complete one episode) and width (i.e.\ the number of actions that an agent can choose from). The environment provides observations as vectors (one-hot vectors suitable for tabular methods) or matrices (2D images suitable for approximate methods). The proposed environment is named the \emph{Configurable Tree graph}, or CT-graph. Table \ref{tab.1} reports a list of properties of the CT-graph that were designed to address the limitations of video game benchmarks. The properties of the CT-graph make this environment suitable to assess the following learning properties of RL algorithms:
\begin{itemize}
\itemsep -3pt
\item learning with variable and measurable degrees of partial observability;
\item learning action sequences of adjustable length and increasing memory requirements;
\item learning with adjustable and measurable sparsity of rewards;
\item learning multiple tasks and testing speed of adaptation (lifelong learning scenarios);
\item learning multiple tasks where the knowledge of task similarity is a required metrics (meta-learning or multi-task learning);
\item learning hierarchical knowledge representation and skill-reuse for fast adaptation to dynamics rewards (lifelong learning scenarios);
\item testing attention mechanisms to identify key states from noise or confounding states;
\item testing meta-learning approaches for optimised exploration policies;
\item learning a model of the environment;
\item learning a combination of innate and learned knowledge to cope with invariant and variant aspects of the environment.
\end{itemize}

Novel reinforcement learning algorithms that implement lifelong learning, incremental learning and optimal adaptation abilities will need to demonstrate such set of skills and properties.

The rest of the paper is organised as follows. Section \ref{CT-graph} presents the definition and main aspects of the CT-graph. Section \ref{sec:learningChallenges} ``Learning challenges'' illustrates examples of various CT-graph configurations. A discussion section (\ref{sec:discussion}) presents concepts that have inspired the idea of the CT-graph and similar benchmarks.

\section{The configurable tree graph (CT-graph)}
\label{CT-graph}

\begin{figure}
\begin{center}
\includegraphics[width=0.9\textwidth]{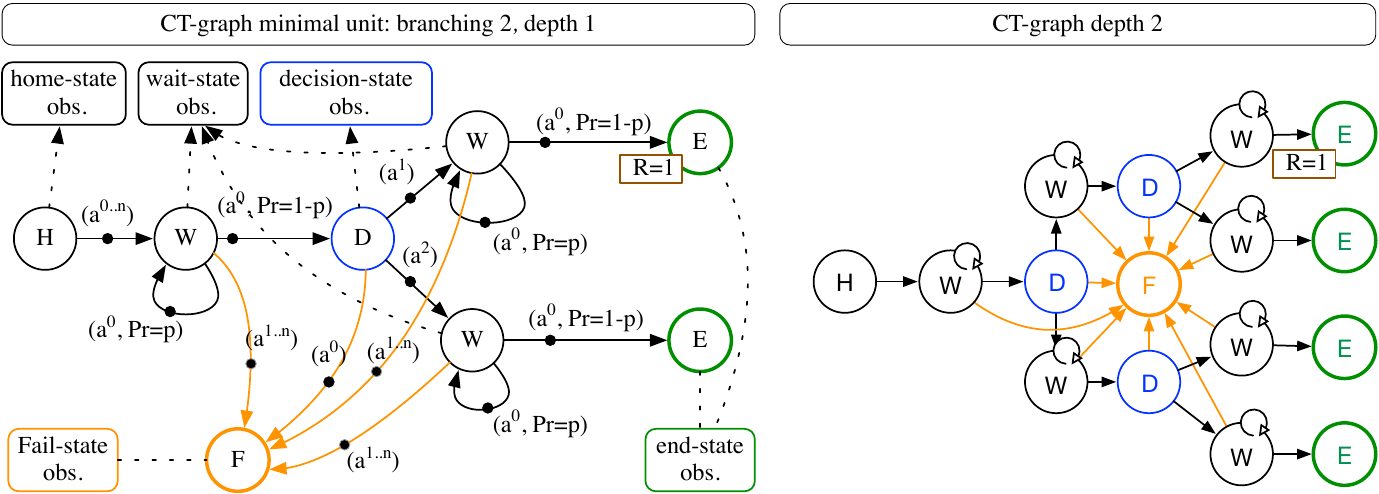}
\caption{\small CT-graph illustrations (Left): Minimal CT-graph unit. The transition graph of the smallest possible graph with $\langle b=2,d=1\rangle$ is shown. (Right): The CT-graph unit on the left is combined to create a larger graph with $d=2$. While these two examples are simple RL problems, partial observability, the wait probability $p$ and the ability to extend the graph to arbitrary depth, allows for the creation of arbitrarily complex and difficult problems.}
\label{fig:MDP}
\end{center}
\end{figure}

The CT-graph represents a family of tree graphs. The objective is to create a learning problem for an agent that learns an optimal sequence of stimuli and actions that maximise the reward over time. An optimal sequence is a task and is defined as the sequence to reach one particular leaf node in the graph. Configurable parameters include the graph's depth, the branching factor of the tree graph, the reward function, and others, making the CT-graph a large family of problems with different sizes and complexity. The unit structure of the graph, which can be repeated to obtain an  arbitrary depth, is illustrated in Fig.\ \ref{fig:MDP}(Left).



One execution of a CT-graph episode is defined as the unrolling of a sequence of stimuli from the set $\mathcal{O}$ and actions from the set $\mathcal{A}$, starting at the home location and ending with either a fail, or at an end state (graph-end) (Fig. \ref{fig:MDP}).

The nodes in the CT-graph belong to five types: \emph{home}, \emph{wait}, \emph{decision}, \emph{graph-end}, and \emph{fail}. Thus, the set $\mathcal{S}$ of all states is the union of the subsets $\mathcal{S}^\mathrm{H} \cup \mathcal{S}^\mathrm{W} \cup \mathcal{S}^{\mathrm{D}} \cup \mathcal{S}^\mathrm{E} \cup \mathcal{S}^\mathrm{F}$. $\mathcal{S}^\mathrm{E}$ and $\mathcal{S}^\mathrm{F}$ are terminal states.
\begin{itemize}
\item \textbf{Home state}: the starting state of each episode and root of the tree graph. Any action leads to the first wait state.\vspace{-0.2cm}

\item \textbf{Decision state}: a state in which the graph forks into $b$ branches, requiring the agent to make a choice out of the $b$ options (or branches) of the graph by selecting one of $b$ different actions in the subset $\mathcal{A}^{\mathrm{D}} =\{A^1,..,A^b\}$. The agent will transition to a fail state if it selects $A^0$.\vspace{-0.2cm}

\item \textbf{Wait state}: a state in which only action $A^0$ leads the agent to the next decision state in the graph with probability $p-1$ and leaves the agent in the same wait state with probability $p$. Other actions lead to the fail state, terminating the episode. Wait states are located before and after decision states. 
\vspace{-0.2cm}

\item \textbf{End state}: a leaf node in the tree graph. It is visited by an agent after having traversed the entire depth of the graph. There are $b^d$  end states. \vspace{-0.2cm}

\item \textbf{Fail state}: if $A^0$ is taken in a decision state or $\neg A^0$ is taken in a wait state, the terminal state \emph{fail} is reached, which returns the agent to the home state.\vspace{-0.2cm}
\end{itemize}

\paragraph{Observations ($\mathcal{O}$).} At each step $t$, the environment provides the agent with an observation $O_t$, which can be a vector (1D setting for standard RL) or an image of dimension $r \times r$ (2D setting for deep RL). A stochastic process $X_{(t,s)}$  maps states in $\mathcal{S}$ to observations in $\mathcal{O}$ and determines the level of observability. The set $\mathcal{O}$ is also divided in five subsets corresponding to those in $\mathcal{S}$: $\mathcal{O} = \mathcal{O}^\mathrm{H} \cup \mathcal{O}^{W} \cup \mathcal{O}^\mathrm{D} \cup \mathcal{O}^\mathrm{E} \cup \mathcal{O}^\mathrm{C}$.
\paragraph{Rewards.} The reward provided by the environment can be found only at an end state. Changing the location of the reward has the effect of changing the reward function only partially and hierarchically: navigation actions to the end states remain constant but the sequence of actions at decision states changes. 


\vspace{0.4cm}
In summary, a CT-graph is defined by the tuple
 \begin{equation}
 \langle b,d,\mathcal{O},\{X_{(t,s)}\},p,g\rangle
 \label{eq:mazetuple}
 \end{equation}
 where
 $b \in \mathbb{N}, b \geq 2$ is an integer greater or equal to 2 that defines the branching factor, i.e., how many choices are available at a decision state. $d \in \mathbb{N}_1$ is an integer greater or equal to 1 that defines the depth of the graph, or the number of decision states between the home and a graph-end. 
 $p$ is the probability to remain in a wait state, resulting in an expected duration of delays in between decision states equal to  1/(p-1). $g$ is the reward function that returns a reward. 
The set $\mathcal{S}$ of states and the set $\mathcal{A}$ of actions are implicitly defined by the tuple \ref{eq:mazetuple}. ~


The execution of a CT-graph experiment typically involves the execution of many episodes, or trials. At each time step t, the observation depends on the action $s_{t-1}$ fed to the environment, on the state of the MDP that is given by tracking the node visited by the agent in the tree-graph, and on the stochastic process $X_{(s,t)}$ that maps states to observations. 

\begin{figure}
\begin{center}
\includegraphics[width=0.8\textwidth]{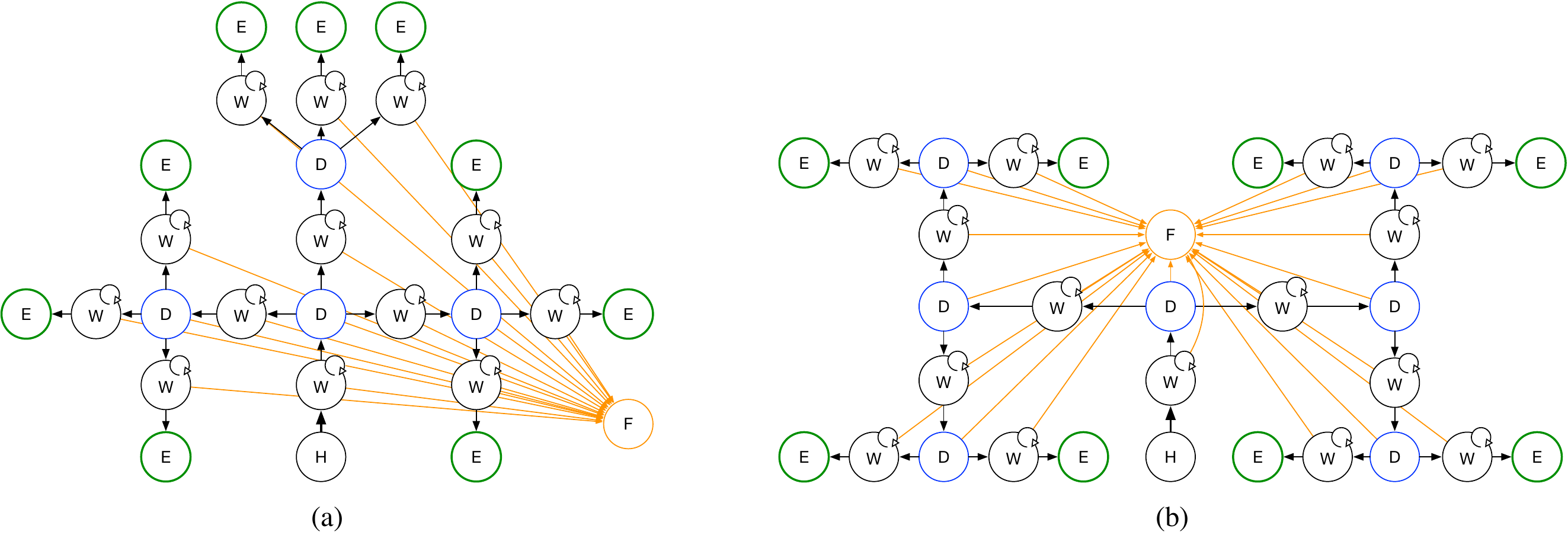}
\caption{\small Examples of CT-graphs. (a) A depth 2 graph with branching 3 $\langle b=3,d=2\rangle$ . (b) A depth 3 graph with branching 2 $\langle b=2,d=3\rangle$ . The colour scheme indicates the different types of states. Small recursive arrows in the wait states indicate that the agent escapes those states with a probability of $1-p$.}
\label{fig:ctD2D3}
\end{center}
\end{figure}

Fig.\ \ref{fig:ctD2D3} illustrates two graphs with depth 2 and depth 3.

\subsection{Distal, sparse, and dynamic rewards}
\label{sec:distal_sparse_dynamic}
A CT-graph is initialised with a sequence of \emph{optimal} decision actions $\mathcal{A}^{\mathrm{D}^\star} \doteq \langle A_1,A_2,...,A_d\rangle $ from the set $\mathcal{A}^\mathrm{D} \doteq \mathcal{A} - A^0$, where $\langle A_1,A_2,...,A_d\rangle$ is a random sequence of integers $A \in \mathbb{N}\,|\, 1\geq a \geq b$ and ($\lvert \mathcal{A}^{\mathrm{D}^\star} \rvert = d$). This sequence defines the location of the reward in the graph. Note that $\mathcal{A}^{\mathrm{D}^\star}$ is the optimal sequence of decision actions at the graph branching points (decision states), but not the complete optimal control sequence that includes a variable number of wait actions during execution.   

\subsection{Hierarchy of policies}
\label{sec:policies}
An agent in the CT-graph may apply policies with the following characteristics and effects:
\begin{itemize}\vspace{-0.2cm}
\item \textbf{Random policy}: The agent performs random actions. This leads to frequent visits to the fail state. Thus, the agent is unlikely to visit deep states, and therefore ever to see a reward. \vspace{-0.2cm}
\item \textbf{Navigation policy}: The agent performs actions to avoid the fail state and thus is able to traverse the graph in its full depth. Such a policy increases significantly the chances to get a reward.\vspace{-0.2cm}
\item \textbf{Optimal policy}: The agent is able to navigate the full graph and applies an optimal policy to maximise the reward by executing the $A^0$ action at wait states and the optimal sequence $\mathcal{A}^{\mathrm{D}^{\star}}$ at decision states.\vspace{-0.2cm}
\end{itemize}

One can interpret the navigation policy as the ability of moving through the graph, which is hard to achieve with a random policy. The optimal policy is the ability to both move through the graph, and choose the specific trajectory that leads to the reward.

\subsubsection{Distal reward ($\rho$)} A reward collected at the end of the graph is distal because is a consequence of a (possibly long) sequence of actions. 
Given this property of the CT-graph, we consider the overall length of an episode as a measure of how distal the final reward is from the graph-home. The number of steps to navigate a graph, $\rho$, is a stochastic value whose mean is given by
\begin{equation}
\mathbb{E}[\rho] = (1-p)^{-1}(d+1) + d + 1\quad.
\label{eq:Erho}
\end{equation}

The CT-graph parameters can be set to obtain a large $\mathbb{E}(\rho)$. For example, a relatively small graph with $d=2$ and $b=2$ can have a large $\mathbb{E}(\rho)$ by setting $p=1-10^{-2}$. In such a case, $\mathbb{E}(\rho) = 100 \cdot (2 + 1) + (2 + 1) = 303$, i.e., there are on average 303 steps between the start of the episode and the reward at the end of the graph.

\subsubsection{Sparsity of rewards}
\label{sec:sparsity}

Rewards are sparse if the probability of an agent to reach a reward is low. If the reward function $g$ provides one reward at one unique graph-end, to obtain it, the agent needs to select $A^0$ in a wait state and the appropriate decision action $A \in \mathcal{A}^{D}$ when at a decision state. The CT-graph allows for computing the following probabilities: \begin{itemize}\vspace{-3pt}
    \item $P_R$: the probability of reaching the rewarding end state with a random policy;\vspace{-3pt}
    \item $P_{E}$: the probability of reaching any end state with a random policy; \vspace{-3pt}
    \item $P_{RNP}$: the probability of reaching the rewarding end state while employing a navigation policy (that avoids the fail state).\vspace{-3pt}
\end{itemize}

Given that the parameter $p$ results in a variable length of an episode, the probability of reaching a reward with a random policy, $P_R$, involves computing the probabilties of all possible trajectories. Thus,
\begin{eqnarray}
	P_{\mathrm{R}} = \sum_{n=0}^{\infty}  \frac{1}{(b+1)^{2d+1+n}} (1+p)^{d+1} p^n \frac{ (n+d)! }{ n! \cdot d! }
	=\nonumber\\= (b+1)\left(\frac{1-p}{(b+1)(b+1-p)}\right)^{d+1}
	\label{eq:randomExpProb}
\end{eqnarray}
For brevity, we omit here the derivation of Eq.\ \ref{eq:randomExpProb} that is reported in the Appendix. 

The probability of reaching any end state, $P_{\mathrm{E}}$ is $P_{\mathrm{R}}$ times the number of end states $b^d$:
\begin{equation}
    P_E = P_R \cdot b^d
    \label{eq:PE}
\end{equation}
$P_{\mathrm{R}}$ and $P_{\mathrm{E}}$ can be very small even for small graphs. E.g., with $b=2,d=2,p=0.9$, the probability of collecting a reward in one episode is $3(0.1/6.3)^3 = 1/83,349$. In other words, applying a random policy, an agent is expected to stumble across a reward approximately once every $83K$ episodes.

If an agent has acquired the skills to navigate the graph to end states (navigation policy, Section \ref{sec:policies}), the probability of reaching the optimal end state is the inverse of the number of end states. We define this value as the reward probability with a navigation policy ($P_{\mathrm{RNP}}$)
\begin{equation}
P_{\mathrm{RNP}} = \frac{1}{b^d}\quad.
\end{equation}
Table \ref{tab:probabilities} provides examples of reward probabilities for different CT-graph configurations. It can be noted that the probability of obtaining a reward by random policy becomes  very small quickly as the parameters $b$, $d$, and $p$ increase. An intelligent exploration policy will find a reward with a probability equal or higher than $P_{RNP}.$

\begin{table}[]
    \centering
    \scriptsize
   \renewcommand{\arraystretch}{1.5}
    \begin{tabular}{|l|c|c|c|c|c|}
    \hline
    CT-graph conf.& $P_R$ & $P_{E}$ & $P_{RNP}$ & states & end states\\
    \hline
         $b=2,d=1,p=0$& $
         3.70\cdot10^{-2}$&$
         7.40\cdot 10^{-2}$ &$2^{-1}$ & 8 & 2\\
         \hline
         $b=2,d=2,p=0.9$&$
         1.20\cdot10^{-5}$&$
         4.78\cdot10^{-5}$&$4^{-1} $& 16 & 4\\
         \hline
         $b=3,d=2,p=0.9$&$
         2.10\cdot10^{-6}$&$
         1.89\cdot10^{-5}$&$9^{-1}$ & 28 &9\\
         \hline
         $b=2,d=4,p=0.9$&$3.02\cdot 10^{-9} $&$4.84\cdot 10^{-8} $&$16^{-1} $&64&16\\
         \hline
         $b=3,d=10,p=0.9$&$3.75\cdot 10^{-23} $&$2.22 \cdot 10^{-18} $&$59049^{-1} $&177148 & 59049\\
         \hline
         $b=2,d=16,p=0.5$ & $3.04 \cdot 10^{-20}$  & $1.99 \cdot 10^{-15}$& $65536^{-1}$& $262144$ & $65536$\\
         \hline
    \end{tabular}
    \caption{\small Examples of reward probabilities per episode, number of states and end states for six different configurations.}
    \label{tab:probabilities}
\end{table}

\subsection{Graph's dimensions}
\label{sec:problem_dimensions}
Given a CT-graph with parameters $\langle b,d\rangle$, the size of the underlying MDP can be computed counting all wait states, decision states, end states, plus two states for the home and fail states:
\begin{equation}
\lvert \mathcal{S} \rvert = f(b,d) = \sum_{x=0}^d{b^x} + \sum_{x=0}^{d-1}{b^x} + b^d + 2 = 2 \sum_{x=0}^d{b^x} + 2\quad.
\label{eq:dim}
\end{equation}
Note that the probability $p$ does not affect the size of the MDP, but is a contributing factor in the complexity of the problem if the set of observations $\mathcal{O}$ is large. 
That is because if $\lvert\mathcal{O}\rvert$ is large, each wait state will randomly manifest itself with a large number of different observations. 

One commonly used approach for RL in POMDP is to map observations to states by deriving a state representation from the history of observations, i.e.\ $S'_t=\langle O_0,O_1,...,O_t\rangle$. However, this is not applicable to a CT-graph if wait states return observations from a large set $\mathcal{O}^{W}$. In fact, due to the stochastic process $X$, the history of observations maps to a large set of equivalent but different states. As the history has variable length according to the stochastic nature of transitions through wait states, the agent cannot infer an MDP from the history, nor from counting the number of steps.

\subsection{Observations}

The CT-graph provides 1D or 2D ($r \times r$) observations. The set of observations $\mathcal{O}$ is divided in five sets that map the sets of states described earlier, $\mathcal{O}^H, \mathcal{O}^W, \mathcal{O}^D, \mathcal{O}^E, \mathcal{O}^F$. The sets' cardinality affects the difficulty of the problem. Since $\mathcal{O}^D$ are observed at decision states, they can be seen as \emph{special} observations, while elements in $\mathcal{O}^W$, provided by wait states can be seen as \emph{distractors}. A special observation requires the agent to choose between multiple actions in $A^T$ in order to continue navigation towards an end state. Special observations are also key to navigate towards the reward. A distractor instead requires the agent to select $a^0$ to continue the navigation towards any end state, and thus, has a simpler relationship with the reward. At each time step t, the observation is a random sample from the specific subset of $\mathcal{O}$ associated with the state in the MDP.


CT-graphs with large observation sets require an RL agent to use experience derived cause-effect relationships or attention models to perform the default $a^0$ action for wait-state observations and identify decision-state observations to perform other actions. This proposition summarises an important aim of the CT-graph environment, i.e., that of reproducing real-world scenarios in which large input-output data streams are not relevant and not causally related to rewards, while a few key stimulus-action pairs are. Human and animal learning has evolved to be robust to such abundance of stimuli and actions and perform a search process that discovers key stimulus-actions pairs.

\subsubsection{Default sets}

The default 1D set is composed of one-hot vectors representing all states. In this configuration $\mathcal{O} = \mathcal{S}$. This is only a testing and debugging feature.

\begin{figure}
    \centering
    \includegraphics[width=\textwidth]{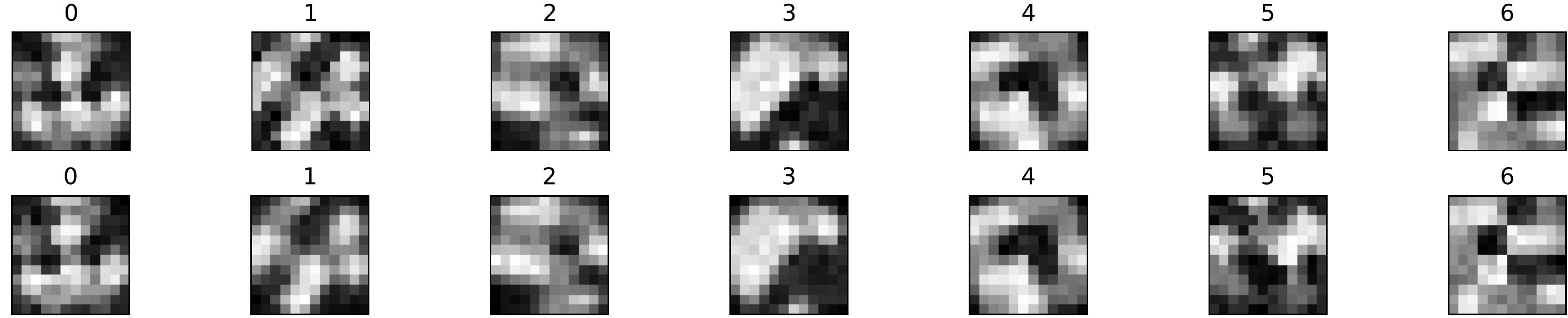}
    \caption{Examples of 7 classes from a set of observations $\mathcal{O}$. The two rows show differences due to noise and rotation each time an observation is fetched.}
    \label{fig:images}
\end{figure}

The default 2D set is built as a synthetic image set with patterns. Each element in $\mathcal{O}$ is generated by upscaling and rotating a random $4 \times 4$ blueprint matrix of $\{0,1,2\}$ elements to result in a $12\times12$ image. The resulting image is then augmented with noise and slightly rotated each time the observation is retrieved and produced by the environment to simulate instances of a class. The upscaling from a $4 \times 4$ random matrix is intended to create an image with local correlations that result into features and patterns. The $12\times12$ observation space allows for a variety of synthetic classes to be generated while maintaining low requirements for the computation and feature extraction process. Fig.\ \ref{fig:images} shows 7 elements from a set $\mathcal{O}$.

The $4\times4$ random blueprint matrix that contains values in the set \{0,1,2\} can generate $3^{16}$ different images. Adding augmentation with two rotations of 30 and 60 degrees, the total space of images is $3\cdot 3^{16}=3^{17}$ (approximately 129 million). In short, the set $\mathcal{O}$ can be built drawing from a large space of different images with patterns that require approximate methods. 

\subsubsection{Alternative observations}

In theory, alternative data sets such as the MNIST, CIFAR-10 or CIFAR-100 could be used instead. These might provide a more real-world set of inputs, but have the limitation of a pre-definite number of classes and instances per class. This could complicate automatic instantiating of large graphs. Additionally, the distances within and between classes is not controllable, introducing perception challenges that cannot be easily measured.  While possible, the current implementation does not include additional data sets.  

\subsection{Reward function}
\label{sec:reward}

Transitions return a reward according to a function $g$. Assume that $A^{D^\circ}$ is the sequence of actions taken by the agent at decision states during one episode. Then
\begin{eqnarray}
g(s_t,a_t) = \left\{ \begin{array}{ll}
\forall \{s_{t+1},a_t\}: s_{t+1} \notin S^E &\Rightarrow g(s,a) = 0\\
\forall \{s_{t+1},a_t\}: s_{t+1} \in S^E & \Rightarrow g(s,a) = c(A^{D^\ast},A^{D^\circ})
\end{array}\right.
\end{eqnarray}
where
\begin{eqnarray}
c(A^{D^\ast},A^{D^\circ}) = \left\{ \begin{array}{ll}1 & \mathrm{if}\quad A^{D^\ast} = A^{D^\circ}\\
0 & \mathrm{otherwise}
\end{array}\right.
\label{eq:needle}
\end{eqnarray}
is a comparison function that returns 1 when the sequence of actions at decision states coincides with the optimal sequence. In other words, the agent receives a reward of 1 for reaching the goal end state, and zero otherwise.  

This setting results in extremely spare rewards as show in Table \ref{tab:probabilities}. Measures can be adopted to simplify the problem by introducing dense rewards, e.g., -1 for reaching the fail state, or, similarly, a small positive value for each step. 

Eq.\ \ref{eq:needle} implies that the reward distribution in the graph is a needle in a haystack. If a new task is created by providing a new sequence $A^{D^\ast}$, an optimal search algorithm will solve the tasks in $\kappa$ episodes, with
\begin{equation}
\mathbb{E}[\kappa] = (b^d+1)/2\quad.
\label{eq:kappa}
\end{equation}
Eq.\ \ref{eq:kappa} provides a lower bound for an optimal exploration strategy that samples all  end states without repetition.

An alternative setting implements a reward gradient across end states. In this case, the function of Eq.\ \ref{eq:needle} becomes 
\begin{equation}
c(A^{D^\ast},A^{D^\circ}) = 1 - \frac{|A^{D^\ast} - A^{D^\circ}| \times [b^d, b^{d-1},..,b^0]^{T}}{b^d-1}\quad,
\label{eq:gradient1}
\end{equation}
where the vector $|A^{D^\ast} - A^{D^\circ}|$ is zero if the goal is reached, and the vector $[b^d, b^{d-1},..,b^0]$ provides weighting parameters for the deviation of $A^{D^\circ}$ from $A^{D^\ast}$. Such a setting provides a gradient across rewards in large graphs such as those in the last two rows of Table \ref{tab:probabilities}.

Finally, rewards can be made stochastic simply by using Eq.\ \ref{eq:gradient1} as the mean in 
\begin{equation}
c^*(A^{D^\ast},A^{D^\circ}) = \mathcal{N}(c(A^{D^\ast},A^{D^\circ}),\sigma)\quad.
\label{eq:stochastic}
\end{equation}
Note that a high standard deviation $\sigma$ in Eq.\ \ref{eq:stochastic} requires a large number of samples to estimate the return. An overview of stochastic n-armed bandit problems in \citet{sutton2018reinforcement} provides examples of the challenges involved in computing such estimates. However, as opposed to n-armed bandit problems that have one single state and multiple actions, the CT-graph requires visiting many states before reaching a graph end where rewards are located. If stochasticity results in negative reward samples, a policy might be ``intimidated'' into not approaching graph ends. Paraphrased with a metaphor, this is equivalent to going out in search of food and ending up being attacked by a predator, which has an immediate larger negative reward than not moving. However, going out and looking for food, although fatal in that particular occasion, is still better than staying at home. 

We would like to point out that while CT-graphs with stochastic rewards are the most realistic types of problems presented here, practically, their complexity is likely to exceed the capabilities of current RL algorithms, unless stochasticity is used for very small graphs (i.e., depth 1 or 2).

\section{Learning challenges}
\label{sec:learningChallenges}

We now present exemplary configurations of the CT-graph that can be used to test RL algorithms for specific learning challenges. Such configurations can be set by modifying the configuration json file, whose entries are reported in Table \ref{tab:json}. The following examples only represent a small set of possible configurations. 
\begin{table}[]
    \centering
    \scriptsize
    \begin{tabular}{|ll|l|l|}
    \hline
        \multicolumn{2}{|l|}{\textbf{Parameter}} & \textbf{Description} & \parbox{0.7in}{\vspace{3pt}\textbf{Common values\vspace{3pt}}}\\
        \hline
         \multicolumn{2}{|l|}{Seed} & \parbox{2.5in}{\vspace{3pt}Seed of the random number generator used for noise and stochastic rewards\vspace{3pt}}& 1;2;...;n\\
         \hline
         \multicolumn{2}{|l|}{graph\_shape} & & \\
         & d & Number of sequential decision states & 1 to 5\\
         & b & Number of branches from a decision state & 2 or 3\\
         & p & \parbox{2.5in}{Probability of remaining in a wait state} & 0; 0.5; 0.9\\
         \hline
        \multicolumn{2}{|l|}{reward} & &\\
         & high\_r & Value of the reward at the goal state& 1\\
         & fail\_r & Value of the reward at the fail state &0; -1\\
         &std\_r & \parbox{2.5in}{Standard deviation on reward sampling}& 0; 0.1\\ 
         \hline
        \multicolumn{2}{|l|}{observations} & &\\
        &MDP\_D&\parbox{2.5in}{If true, each decision state provides a unique obs.\vspace{3pt}}&true, false\\ 
        &MDP\_W &\parbox{2.5in}{If true, each wait state provides a unique obs.\vspace{3pt}}&true, false\\
        &W\_IDs & \parbox{2.5in}{Start and end indices of obs. for wait states\vspace{3pt}}& [$1,\lvert \mathcal{O}^W\lvert$]\\
         &D\_IDs & \parbox{2.5in}{Start and end indices of obs. for decision states\vspace{3pt}}&[$\lvert \mathcal{O}^W\lvert$ + 1,$\lvert \mathcal{O}^D\lvert$]\\
         \hline
          \multicolumn{2}{|l|}{image set} & &\\
         & seed & Specific seed for image data set& 1;2;...;n\\
         & 1D & Use 1D one-hot vectors as obs.&false, true\\
         &nr\_images & \parbox{2.5in}{Number of images to be created in the data set\vspace{3pt}}& 5 or higher\\ 
        &noise on read & \parbox{2.5in}{Noise on each pixel when an obs.\ is read\vspace{3pt}}& \\
        &rotation on read & \parbox{2.5in}{Maximum random rotation when an obs.\ is read\vspace{3pt}}& 0 deg; 5 deg\\
\hline
    \end{tabular}
    \caption{List of parameters available in the json configuration  file.}
    \label{tab:json}
\end{table}

\subsection{Fully observable graph}


The CT-graph can be configured to be fully observable. This is useful mainly for debugging and testing basic RL algorithms. Full observability in a CT-graph can be obtained replacing $X_{s,t}$ with an injective function $f$ to map each state to an observation. In other words, each different decision state and each different wait state have a \emph{reserved} observation from the set $\mathcal{O}$. By setting MDP\_D = true and MDP\_W = true, each state in the graph has unique observations, and thus becomes fully observable.  

Configuration \textbf{CT-FO-B1}: Fully observable, baseline 1.

\noindent\textbf{Parameters:} d=2, b=2, p=0, MDP\_D=true, MDP\_W=true.

\noindent\textbf{Properties:} A small graph with only 3 decision states, 7 wait states, 4 end states. The number of steps per episode is $\rho = 6$. The probability per episode of scoring a high-reward with a random policy is $P_R$ = $1/2^5 = 1/32$. 

\noindent\textbf{Suitability:} Basic checks and debugging. 

Configuration \textbf{CT-FO-B2}: Fully observable, baseline 2.

\noindent\textbf{Parameters:} d=3, b=2, p=0.5, MDP\_D=true, MDP\_W=true.

\noindent\textbf{Properties:} A larger graph with 7 decision states, 15 wait states, 8 end states. The number of steps per episode is $\mathbb{E}[\rho] = 11$. $P_R = 5.93 \cdot 10^{-5}$, or 1 reward in 16875 episodes.

\noindent\textbf{Suitability:} Testing standard RL algorithms with sparse rewards. 

\subsection{Learning with variable degrees of observability}

\subsubsection{The surjective CT-graph}
In the surjective CT-graph, partial observability is introduced by using a surjective function $f$ instead of $X_{(s,t)}$: only five observations corresponding to each type of the five types of state are used. 

\begin{figure}
\begin{center}
\includegraphics[width=0.95\textwidth]{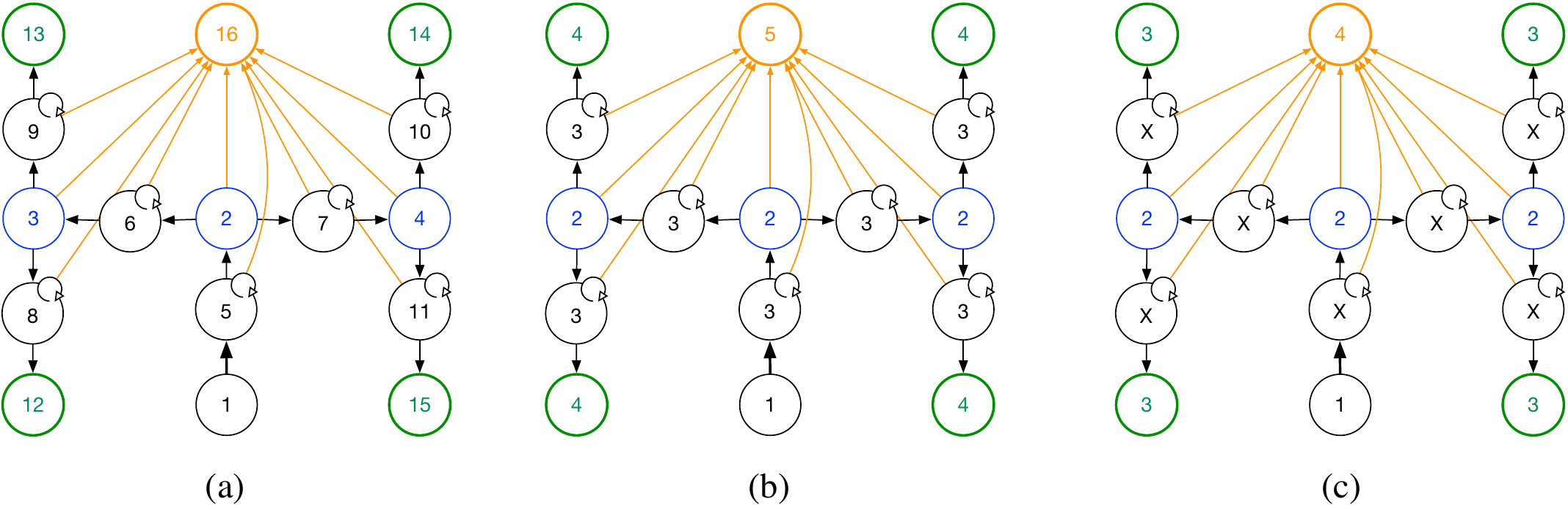}
\caption{Fully observable, surjective and confounding CT-graphs. The numbers inside the nodes are the class ID of the observations. (a) Fully observable: each state in the MDP has a unique class in the observation set. (b) Surjective graph: each state type provides one observation. (c) Confounding graph: the wait states provide stochastic observations from a large set $\mathcal{O}^W$.}
\label{fig:surjective}
\end{center}
\end{figure}


The surjective graph is a POMDP that models navigation environments in which the same visual inputs repeat in different states of the MDP. Such a graph cannot be solved by vanilla RL if observations are treated as states. 

Configuration \textbf{CT-SU-B1}: Surjective, baseline 1.

\noindent\textbf{Parameters:} d=2, b=2, p=0, MDP\_D=false, MDP\_W=false, D\_IDs=[2,2], W\_IDs=[3,3]

\noindent\textbf{Properties:} A small graph with only 3 decision states, 7 wait states, 4 graph ends. The number of steps per episode is $\rho = 6$. The probability per episode of scoring a high-reward with a random policy is $P_R$ = $1/2^5 = 1/32$. However, all decision states provide the same observation, and similarly all wait states. 

\noindent\textbf{Suitability:} Testing RL for POMDP with either memory or belief systems. Algorithms without memory or belief system will be unable to reach particular graph ends for which different actions are required at different decision states that provide the same observation.

\subsubsection{The confounding CT-graph}

Employing the stochastic process $X_{(s,t)}$ to generate observations results in a higher level of non-observability and more challenges for the learning agent. In the case of the \emph{confounding} CT-graph, we assume that all decision states provide the same observation from the set $\mathcal{O}^D$ of cardinality one, and all wait states provide random observations from a large set $\mathcal{O}^W$ with $\lvert \mathcal{O}^W \rvert >> 1$.


The challenge with the confounding CT-graph is that the agent needs to learn to ignore all seemingly random stimuli and perform $a^0$ at wait states, while learning decision states where the correct decision action is required. 

The confounding graph is the most interesting configuration setting in which cause-effect relationships have to be extracted and separated from random stimuli. Approaches such as attention, associative learning and neurmodulation are likely to provide an advantage in this case.

Configuration \textbf{CT-CO-B1}: Confounding, baseline 1.

\noindent\textbf{Parameters:} d=2, b=2, p=0, MDP\_D=false, MDP\_W=false,  D\_IDs=[2,2], W\_IDs=[3,102]

\noindent\textbf{Properties:} A small graph with only 3 decision states, 7 wait states, 4 graph ends. The number of steps per episode is $\rho = 6$. The probability per episode of scoring a high-reward with a random policy is $P_R$ = $1/2^5 = 1/32$. However, all decision states provide the same observation. Wait states provide any observation from a set of 100 images.  

\noindent\textbf{Suitability:} Testing RL for POMDP with either memory or belief systems, and attention systems, or particular abilities to learn particular features in the input stream and ignore others.

\subsubsection{Fully stochastic-observation graph}

If the cardinality of all subsets is greater than one, i.e., $\lvert O^H \rvert  = \lvert O^D\lvert = \lvert O^W \rvert = \lvert O^E \rvert = \lvert O^F \rvert> 1$, the challenge increases because the agent needs to learn an explicit association between groups of classes and states of the POMDP. While an agent in the confounding graph could learn to identify decision states, and ignore all other observations, in a fully stochastic-observation graph, the agent is required to derive the association of each image in $O$ with the specific subset. In practice, this is an extremely hard problem  that links to real-world scenarios only if each subset of $O$ share common features. In other words, if observations within one subset are more similar to each other than observations among different sets, it is possible for an RL agent to learn what makes a wait observation different from a decision state observation.






\subsection{Learning with distal and sparse rewards}

The length of the graph, expressed by $\rho$ in Eq.\ \ref{eq:Erho}, determines a distal measure of the rewards, or in other words, the length of the episode. 

In all previous cases with a deterministic $\rho$, an agent can learn to reach a specific goal state by ignoring all inputs, and simply applying the unique and optimal sequence. If the probability of staying in a wait state, $p$, is set to be greater than 0, $\rho$ becomes stochastic. In this case, the agent is required to detect the difference between a decision state and a wait state to perform an optimal sequence of actions. Additionally, a high value of $\rho$ increases the sparsity of the reward and decreases the probability of finding a reward by a random policy. 

Configuration \textbf{CT-POSR-B1}: Partially observable sparse reward, baseline 1.

\noindent\textbf{Parameters:} d=2, b=2, p=0.5, MDP\_D=false, MDP\_W=false,  D\_IDs=[2,2], W\_IDs=[3,3]

\noindent\textbf{Properties:} A small graph with only 3 decision states, 7 wait states, 4 end states. The number of steps per episode is stochastic and given by Eq.\ \ref{eq:Erho}, which in this case is $\mathbb{E}[\rho] = 9$. The probability per episode of scoring a high-reward with a random policy is $P_R = 0.00088 = 8.88 \cdot 10^{-4}$ (see Eq.\ \ref{eq:randomExpProb}). All decision states provide the same observation. All wait states provide the same observation.  

\noindent\textbf{Suitability:} Testing RL for POMDP with either memory or belief systems on problems with distal and sparse rewards and a variable duration of episodes.


\subsection{Lifelong learning across multiple tasks}

There are multiple ways to generate multiple tasks, which can then be assembled in a curriculum in which tasks can be learned sequentially according to a lifelong or continual learning protocol. The CT-graph allows for variations across three different domains: (1) variation of reward function; (2) variation of input distributions; (3) variation of the MDP; 

\textbf{Variation of the reward function}. For a given CT-graph with $n$ end states, it is possible to create $n$ tasks, each of which has the high reward at a different end state. The input distribution and the structure of the MDP remain unchanged. This means that different tasks have a different sequence of optimal actions, or policy. Thus, such tasks are adversarial, meaning that one single function cannot learn more than one such a task. 

\textbf{Variation of the input distribution}. Multiple tasks can be generated by changing some or all classes in the observation sets. This can be done by selecting different image IDs, or simply by creating a new data set with different seeds. 

\textbf{Variation of the MDP}. Two CT-graphs that have different shapes, e.g., a depth 2 and a depth 3, will have different MDP structures. 

\subsubsection{Exploiting task similarities for lifelong learning}

If two CT-graphs have different shapes, different input distributions and different reward functions, they can be used in a LL curriculum to test particular learning properties. In particular, their lack of similarities can be exploited to test an LRL to learn new and uncorrelated tasks without suffering from catastrophic forgetting. However, such a curriculum will not test the ability to exploit old and new knowledge in backwards and forward transfer metrics \citep{BAKER2023}.

A more interesting curriculum can be built by creating tasks that share similarities. One such example are two surjective graphs that share the same inputs and MDP, but have different reward functions. Using the example above CT-SU-B1, it is possible to create a curriculum of 4 tasks, each with the reward at at different end state. Solving all tasks requires learning different functions. However, the policy required at wait states is the same. This similarity can be exploited by LRL algorithms to accelerate learning across tasks. 

We list here some practical ways to generate LL curricula with task similarities.
\begin{itemize}
    \item Surjective graph with different reward locations. The similarities are the policy at the wait states.
    \item Surjective graph with similar reward locations. The similarities are both in the policy at the wait states and at some of the decision states.
    \item Graphs with growing depth. A depth 3 graph will contain a depth 2 graph, meaning that they share similarities. Additionally, similarities can be created if the shallower graph has a goal location along the trajectory of the goal location in the deeper graph. 
\end{itemize}

\subsection{Coping with variant and invariant features}

Assume that a LRL algorithm is given a large set of $n$ different graphs. All such graphs, while different, share some similarities.
The similarities represent invariant features across all problems. Therefore, an ideal learner could implement a policy that is composed of a fixed part to deal with invariant feature, and a learnable policy to adapt to variant features. Such a distinction introduces a hierarchy of knowledge that can be exploited by meta-RL algorithms, evolutionary algorithms that evolve both inborn knowledge and learning strategies \citep{soltoggio2018born}, and more in general LRL algorithms that can exploit task similarities to accelerate learning. 

%

\subsection{Code}

The CT-graph is implemented as an OpenAI gym environment. The code is available at \url{https://github.com/soltoggio/CT-graph}.







\section{Discussion}
\label{sec:discussion}

The inspiration that lead to the development of the CT-graph is discussed in the following section. A brief overview of known performance of RL algorithms is provided.

\subsection{Inspiration}
\label{background}
The CT-graph is an abstract and generalised problem formulation that draws elements from a range of concepts such as sequence learning \citep{clegg1998sequence}, backup diagrams  \citep{sutton2018reinforcement}, the well known T-maze environment used in animal behavior  \citep{olton1979mazes,wenk1998assessment} and computational studies \citep{soltoggio2008evolutionary}, and the problem of learning from rich and confounding stimuli-actions sequences \citep{izhikevich2007solving, soltoggio2013solving, soltoggio2013rare}.

Decision problems can be seen as a sequence of stimuli and actions that lead to a desired state. A practical example is a T-maze environment in which an animal, typically a rat, starts at the bottom of the maze and walks forward until it reaches a turning point. On either sides there is a reward or a punishment. An animal such as a rat will learn which turning direction to choose to maximise reward and reduce punishment. Multiple T-junctions can be added to increase the length of the sequence of actions to memorise. The sequence of correct turning actions to reach a reward is an arbitrary long sequence of optimal actions. As such, the problem can be formulated as a reinforcement learning problem. If turning points and corridors look similar, observations do not maps to states in a MDP, and a memory system is required to perform navigation. 

\begin{figure}
\begin{center}
\includegraphics[width=0.7\textwidth]{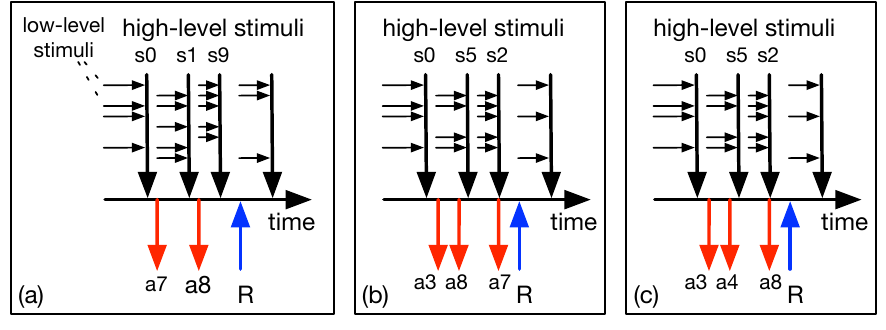}
\end{center}
\caption{\small Example of ambiguous cause-effect information stream (problem formulation derived from \citet{izhikevich2007solving} and \citet{soltoggio2013solving}. In the figure, the distinction between low and high-level stimuli represents the fact that cause-effect relationships are likely to exist at high levels of representations. (a) A series of stimuli and actions precede the delivery of a reward. From the observation of these series, any stimuli, any actions, or any combinations of stimuli and actions that precede the reward could be the cause. If a second sequence is observed, panel (b), we can restrict the possible causes of the reward to {s0, a8, a7, s0+a8, s0+a7}. Finally, in this particular example, a third observation in panel (c) allows us to determined that the reward is caused by s0+a8. Irrelevant or confounding stimuli are implemented in the CT-graph with wait states and large sets of observations.}
\label{fig:ambiguous}
\end{figure}

Additionally, while executing a sequence of stimuli-informed actions, distracting or random stimuli and actions may provide spurious correlations with rewards. For example, driving from A to B might lead a driver to observe identical cues, e.g., speed limit signs or cars parked on the side of the road, that do not reveal the state along the path because they repeat similarly at different locations. Thus, finding the optimal sequence of actions requires the ability to discount irrelevant information, and extract and focus on a subset of stimulus-action sequences that lead to a reward. This condition can be abstracted as in the information stream presented in Fig. \ref{fig:ambiguous}. If there are cause-effect relationships, but other stimuli and actions occur simultaneously, learning of true cause-effect relationships require the observation of many occurrences of the reward to extract the unique causing factors. The challenge increases as the delay among stimuli, actions, and reward increase because it also increase the number of intervening confounding stimuli that are not causally related to reaching a reward. 


\subsection{Known performance of RL algorithms on the CT-graph}

The range of SoTA RL algorithms that can address the many learning challenges in the CT-graph is too large to provide comprehensive tests as part of this paper. We choose instead to list the papers that have used the CT-graph so far and summarise the main results. 

In \citet{ben2020evolving}, a surjective graph (POMDP) with depth 2 and 3 was used to learn multiple reward functions sequentially (different goal locations). Meta-RL algorithms were compared with an evolutionary approach. Due to the memory requirements, partial observability and adversarial tasks, CAVIA \citep{zintgraf2019fast}, MAML \citep{finn2017model} and RL$^2$\citep{duan2016rl} failed to solve all tasks. The neuroevolution approach instead evolved memory units triggered by particular observations, which lead to solve the tasks. A similar depth 2 graph was used in \citet{dick2020detecting} to test the ability of a statistical approach to detect task changes. The similar graphs had a single variation in the transition matrix, which made the two environments very similar and therefore difficult to detect. However, the performance of RL algorithms was not assessed. 

In \citet{ladosz2021deep}, a confounding CT-graph was used to test a RL architecture that combined backpropagation with an associative Hebbian neural unit. The most complex benchmark had a high $p=0.9$, which led to long trajectories due to many cycles in the wait states, as well as a large number of observations (500). The setting aimed to reproduce a real-world condition in which few relevant task-cues are to be discovered among a large amount of diverse irrelevant observations. While the proposed algorithm was devised to perform well under such challenging conditions, several baselines, including DQN \citep{mnih2013playing}, QRDQN+LSTM \citep{hausknecht2015deep}, REINFORCE \citep{williams1992simple}, A2C \citep{mnih2016asynchronous}, AMRL \citep{beck2019amrl} and Backpropanine \citep{miconi2018differentiable} performed poorly or failed completely on the most complex graphs.  

In \citet{ben2022context}, a meta-RL method that used a neuromodulatory approach attempted to solve full-MDP graphs of depth 2, 3 and 4 with variying reward functions. CAVIA \citep{zintgraf2019fast} and PEARL \citep{rakelly2019efficient} were shown to perform well with the addition of neuromodulation in tests where multiple tasks were instantiated with different goal locations. 

In \citet{ben2022lifelong}, a CT-graph of depth 5 was solved for the first time. The graph was MDP for the decision states (MDP\_D=true) and POMDP for the wait states (MDP\_W=false), which led to graphs that can be solved without memory and share similar wait states.  The graph with depth 5 had a very small reward probability of $5.6 \cdot 10^{-6}$ (one reward every 177,147 episodes with a random policy). The novel algorithm investigated the potential of modulating masks in lifelong RL. The depth 5 graph was solved only by a system that first learned to solve depth 2, 3 and 4, in such order, before tackling depth 5. A linear combination of masks was used to exploit previously learned knowledge and apply it to the most challenging task. 

\subsection{Limitations}

The CT-graph may not be the most appropriate benchmark for all RL problems as some distinctive features lead to both advantages and disadvantages.  

While input spaces can be customised to larger and more complex images, the standard synthetic set is designed for speed and for having equidistant classes. Expanding the CT-graph with other data sets is not immediate, and the new benchmark will have different properties, making comparisons difficult. A related limitation is that the current input set is not designed to have correlations among observations: in a full MDP setting, each observation is distinct. As a result, the power of approximate methods may not be fully tested. Setting a high level of noise when fetching observations will provide partial compensation for the problem. Another solution is to alter the image generation process to introduce correlations or similarities among particular observations, e.g., decision state observations, but again, that is not part of the current implementation.

The discrete action space makes the CT-graph less suitable to test algorithms for continuous output. 

The nature of a tree graph is that it expands exponentially as the agent moves away from the start node. Each step in the CT-graph represents an action that cannot be undone, i.e., the agent cannot go back along the tree. As a result, there is only one optimal trajectory, which may not be the case for other RL environments where multiple optimal trajectories can exist.  

Finally, while the many configurations offer a large range of problems, they also imply that the CT-graph is not a single reference benchmark. Different algorithms can be compared only if they are tested on the same configurations. 

\section{Conclusion}
This paper describes a set of mathematically formulated RL problems that, despite their apparent simplicity, can be used to create different and hard challenges for RL algorithms. The problems are inspired by real-world scenarios where partial observability, confounding stimuli, distal rewards and multiple tasks affect the efficacy of RL algorithms. While this benchmark does not use appealing or catching visual inputs as 3D FPV environments, it has the advantage to allow for precisely defined levels of complexity. In particular, the CT-graph can be set to provide varying levels of observability, measurable sparsity of the reward, measurable size of the MDP, and others.  The depth of the graph can be easily configured to obtain extremely sparse rewards and large MDPs that render SoTA RL algorithms ineffective. Therefore, the CT-graph is particularly suitable when specific learning properties of a RL algorithm need to be rigorously assessed against precisely defined learning challenges. Finally, this benchmark is suitable to test lifelong learning capabilities due to the ability to generate large number of tasks with various degrees of similarities.

\subsection*{Acknowledgement}

This material is based upon work supported by the United States Air Force Research Laboratory (AFRL) and Defense Advanced Research Projects Agency (DARPA) under Contract No.\, FA8750-18-C-0103 (Lifelong Learning Machines) and Contract No.\, HR00112190132 (Shared Experience Lifelong Learning).

\section*{Appendix}
\label{sec:AP1}

\subsection*{Derivation of the probability of a reward}

We show here how to compute the probability of obtaining a reward while applying a random policy. Firstly, we need to compute the probability of an episode of being of a given length $\rho$. The probability of $\rho$ taking a particular value $l$ can be found in the following manner. Let a \emph{wait transition} be a transition from a wait state back to the same wait state, and an \emph{escape transition} be a transition from a wait state to the next decision or end state. Firstly, find the probability of the graph having, for example, all transitions from wait state to wait state occur before the first decision state, and the agent avoiding any such transitions at all following wait states. The minimum possible value of $\rho$ is given by $2(d+1)$, so the number of wait state to wait state transitions $n$ required to make $\rho = l$ can be set as $n = l-2(d+1)$. The probability of making a wait state to wait state transition is $p$, and the probability of escaping a wait state is given by $1-p$. The probability of making $n$ wait transitions followed by only escape transitions until the end of the graph is $p^{n} \times (1-p)^{d+1}$. Multiply this by the number of ways the $n$ wait transitions can be arranged between the escape transitions. The only restrictions are that the number of wait transitions must be $n$, the number of escape transitions must be $d+1$, and the sequence of transitions must end in an escape transition. This means we have to take a combination of $d$ escape transitions from a collection of $n + d$ total transitions. The number of possible combinations is $\frac{ (n+d)! }{ n! \times d! }$.
So, the total probability of $\rho$ taking the value $l = n + 2(d+1)$ is
\begin{equation}
	P(\rho = n + 2(d+1)) = p^{n} \times (1-p)^{d+1} \times \frac{ (n+d)! }{ n! \times d! }\quad.
	\label{specificRho}
\end{equation}
To find the expected probability of the agent reaching one particular graph end through random exploration in one episode 
, we multiply the probability of the agent reaching the graph end at each length $l$ by the probability of the graph being length $l$, and sum these probabilities for all $l$. For $l < 2(d+1)$, the probability $P_(\rho = l) = 0$. So, we can start our summation with $l = 2(d+1)$, which is equivalent to $n = 0$.

Recall that $n$ is the number of times the agent transitions from the wait state back into the wait state, and $b$ is the branching factor, or number of paths available at a decision state. Note that $b+1$ actions can be chosen at any state. Also, recall that at the home state, any action results in a transition to the next state. This means that of all $l$ actions taken, only $l-1$ must be specific. $P_R$ is therefore expressed by:
\begin{eqnarray}
	P_{\mathrm{R}} = \sum_{n=0}^{\infty}  \frac{1}{(b+1)^{2d+1+n}} (1+p)^{d+1} p^n \frac{ (n+d)! }{ n! \cdot d! }
	=\nonumber\\= (b+1)\left(\frac{1-p}{(b+1)(b+1-p)}\right)^{d+1}
 \nonumber
\end{eqnarray}

\bibliographystyle{apalike}
\bibliography{biblio.bib}

\end{document}